\documentclass[a4paper, 10pt, conference]{ieeeconf}
\usepackage{graphicx}
\usepackage{makecell}
\usepackage{cite}
\usepackage{multirow}
\usepackage{multibib}
\usepackage{url}
\usepackage{threeparttable}

\IEEEoverridecommandlockouts                              

\title{\LARGE \bf
Deep Learning based Food Instance Segmentation using Synthetic Data
}

\author{Deokhwan Park$^{1}$, Joosoon Lee$^{1}$, Junseok Lee$^{1}$ and Kyoobin Lee$^{1}$

\thanks{$^{1}$School of Integrated Technology (SIT), Gwangju Institute of Science and Technology (GIST), Republic of Korea. {\tt\small joosoon1111@gist.ac.kr}}

\thanks{© 2021 IEEE. Personal use of this material is permitted. Permission from IEEE must be obtained for all other uses, in any current or future media, including reprinting/republishing this material for advertising or promotional purposes, creating new collective works, for resale or redistribution to servers or lists, or reuse of any copyrighted component of this work in other works.}
}%

\begin{document}

\maketitle
\thispagestyle{empty}
\pagestyle{empty}

\begin{abstract}
In the process of intelligently segmenting foods in images using deep neural networks for diet management, data collection and labeling for network training are very important but labor-intensive tasks. In order to solve the difficulties of data collection and annotations, this paper proposes a food segmentation method applicable to real-world through synthetic data. To perform food segmentation on healthcare robot systems, such as meal assistance robot arm, we generate synthetic data using the open-source 3D graphics software Blender placing multiple objects on meal plate and train Mask R-CNN for instance segmentation. Also, we build a data collection system and verify our segmentation model on real-world food data. As a result, on our real-world dataset, the model trained only synthetic data is available to segment food instances that are not trained with 52.2\% mask AP@all, and improve performance by +6.4\%p after fine-tuning comparing to the model trained from scratch. In addition, we also confirm the possibility and performance improvement on the public dataset for fair analysis. Our code and pre-trained weights are avaliable online at: \url{https://github.com/gist-ailab/Food-Instance-Segmentation}

\end{abstract}

\section{INTRODUCTION}
Some experts predict that 38 percent of adults in the world will be overweight and 20 percent obese by 2030 if the trend continues \cite{kelly2008global}. Due to the increasing obesity rate, the importance of diet management and balanced nutrition intake has recently increased. In particular, services are gradually being developed to automatically calculate and record kinds of food and calories through photos of food to be consumed. The most important technology in this service is food recognition and can be widely used in a variety of service robots, including meal assistance robots, serving robots, and cooking robots.

Because of increasing importance of food-aware tasks, many researchers are working hard on the production of food-aware datasets and the development of food recognition. There are three methods to recognize food: food classification, food detection, and food segmentation. Food classification is a task that matches the type of food in an image through a single image, and many public datasets are also released because it is relatively easier than other tasks  during the data collection and labeling phase. However, in order to determine a more accurate food intake in the diet management service, it is necessary to pinpoint the real food portion within the image. Therefore, food segmentation are more useful in this service than food classification and food detection which provides information of the food types and the position by expressing in a bounding box. Nevertheless, there are three difficulties in food segmentation. First, as shown in Table \ref{table:dblist}, released public datasets available for food segmentation are very scarce compared to food classification public datasets, and most datasets are not publicly available even if released. Second, when producing a dataset personally, there are a tremendous variety of food types and it takes a huge labor cost to labeling. Third, food segmentation is still a challenging task because the variations in shape, volume, texture, color, and composition of food are too large.

To address the presented difficulties, we employed two methods. First, We introduced deep neural network for instance food segmentation. In the early works of food segmentation, multiple food items were recognized mainly through image processing techniques: Normalized Cuts \cite{zhu2014multiple}, Deformable Part Model \cite{matsuda2012recognition}, RANSAC \cite{anthimopoulos2013segmentation}, JSEG \cite{ciocca2016food, matsuda2012recognition}, Grab Cut \cite{fang2018ctada}, and Random Forest \cite{inunganbi2018classification}. In those cases, the sophistication of technique is more important than the acquisition of datasets. Lately, with the introduction of deep learning, deep neural network has eliminated the hassle of image processing by finding food features in the image on its own. There is a study that simultaneously localizes and recognizes foods in images using Fast R-CNN\cite{shimoda2015cnn}. Moreover, there is CNN-based food segmentation using pixel-wise annotation-free data through saliency map estimation\cite{bolanos2016simultaneous}. However, most relevant studies do not distinguish the same type of food in different locations as semantic segmentation, and it is most important to provide sufficient data to allow itself to learn. In that sense, secondly, we generated synthetic data and train these to apply food segmentation in real-world environments, called Sim-to-Real technique. The Sim-to-Real is an efficient technique that is already being studied in robot-related tasks, such as robot simulation \cite{golemo2018sim} and robot control \cite{peng2018sim}, etc. Also, it has the advantage of overcoming environmental simulations or data that are difficult to implement in real-world environments. Using this application, segmentation masks were easily obtained by randomly placing plates and multiple objects in a virtual environment to create a synthetic data describing the situation in which food was contained on the plate in a real world. Using this synthetic data and the Mask R-CNN \cite{8237584} model, which is most commonly used in segmentation tasks. We conduct a class-agnostic food instance segmentation that recognizes that food types are not classified (only classify background and food) but are different. Furthermore, we found the following effects:

\begin{itemize}
  \item Unseen food instance segmentation of first-time tableware and first-time food is possible in real-world environments through random object creation of synthetic data and deep learning
  \item Food segmentation is sufficiently possible in real-world environments when learning using synthetic data only 
  \item After learning with synthetic data, fine-tuning with real-world data improves performance
  \item By distinguishing the same food in different locations, it can be used efficiently in robot fields, such as food picking, which can be utilized later
\end{itemize}

This paper is divided into 4 sections including this introduction section. In section 2, the data production process of synthetic and real-world data, models and parameters used in learning, and evaluation metrics used in performance comparisons are described. In section 3, performance comparison results were described according to the combination of learning data: synthetic data, our real-world data we collected, and public data called UNIMIB2016 \cite{ciocca2016food2}. Finally, in section 4, the conclusions are given.

\begin{table}[]
\renewcommand{\arraystretch}{1.3}
\caption{List of food datasets}
\label{table:dblist}
\begin{tabular}{llc}
\Xhline{3\arrayrulewidth}
Name & Task & Reference \\
\Xhline{3\arrayrulewidth}
Food50 & Classification & \cite{10.5555/1818719.1818816} \\
PFID & Classification & \cite{chen2009pfid} \\
TADA & Classification & \cite{mariappan2009personal} \\
Food85 & Classification & \cite{hoashi2010image} \\
Food50Chen & Classification & \cite{chen2012automatic} \\
UEC FOOD-100 & Detection & \cite{matsuda2012recognition2}\\
Food-101 & Classification & \cite{bossard2014food}\\
UEC FOOD-256 & Detection & \cite{kawano2014automatic}\\
UNICT-FD1200 & Classification & \cite{farinella2016retrieval}\\
VIREO & Classification & \cite{chen2016deep}\\
Food524DB & Classification & \cite{ciocca2017learning}\\
Food475DB & Classification & \cite{ciocca2018cnn}\\
MAFood-121 & Classification & \cite{aguilar2019regularized}\\
ISIA Food-200 & Classification & \cite{min2019ingredient}\\
FoodX-251 & Classification & \cite{kaur2019foodx}\\
ChineseFoodNet & Classification & \cite{chen2017chinesefoodnet}\\
\hline
UNIMIB2015 & Classification and Leftover & \cite{ciocca2015food} \\
Foood201-Segmented & Segmentation & \cite{meyers2015im2calories}\\
UNIMIB2016 & Classification and Segmentation & \cite{ciocca2016food2} \\
SUECFood & Segmentation & \cite{gao2019musefood} \\
Food50Seg & Segmentation & \cite{aslan2020benchmarking}\\
\Xhline{3\arrayrulewidth}
\end{tabular}
\end{table}

\section{METHODS}
We propose a unseen food segmentation method that enables segmentation of untrained foods from real-world images. We used a deep neural network and constructed a synthetic dataset and a real-world dataset for unseen food segmentation using deep learning. For training deep neural network, data is the most important factor. In reality, however, it is quite challenging to build an appropriate dataset for every task. Therefore, we used Sim-to-Real, which learns deep neural networks using synthetic data and applies them to real-world. If we get real food data, a lot of time and expense is needed for data collection and annotation. So, we generated synthetic data using Blander simulator, a computer graphics software, to conserve resource.  Also, we collected real-world data by building our food image acquisition system for verification of unseen real-world food segmentation. 

\subsection{Dataset} 

\textbf{Synthetic Dataset} The use of synthetic data for training and testing deep neural networks has gained in popularity in recent years \cite{back2020segmenting, danielczuk2019segmenting}. The Blender, a 3D computer graphics production software capable of realistic rendering \cite{blender279}, is often used to create synthetic data. Realistic high-quality synthetic data is required for deep neural networks to show high performance for real situations. Therefore, we generated the synthetic data using Blender. 

\begin{figure}[h!]
    \centering
    \includegraphics[scale=0.5]{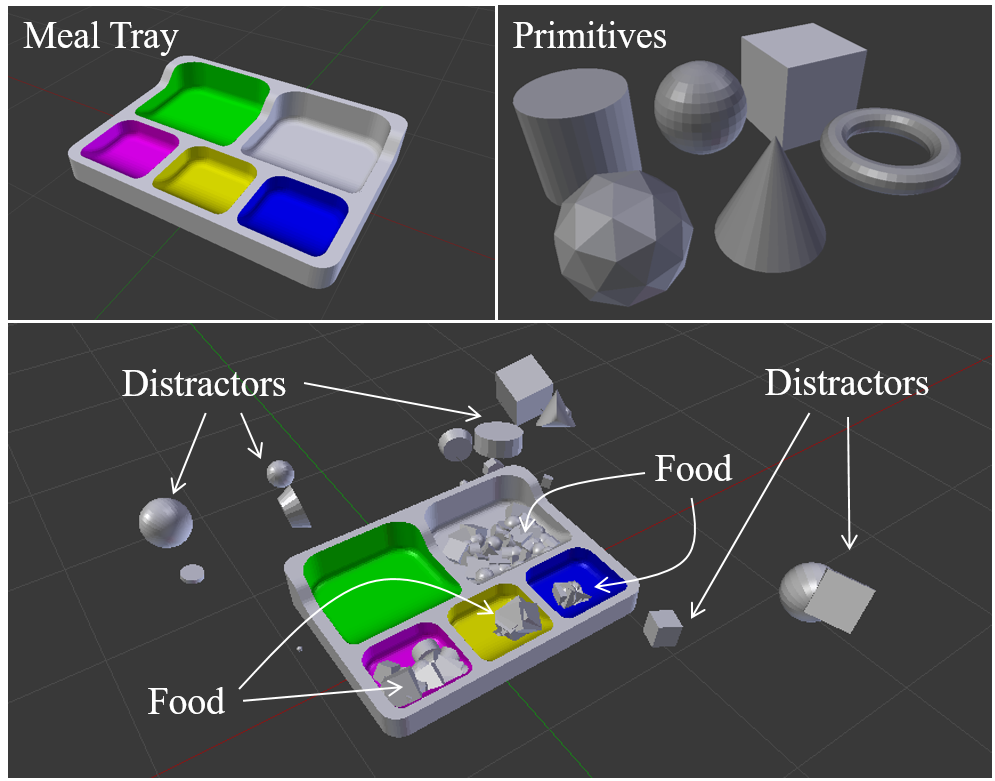}
    \caption{Examples of synthetic dataset}
    \label{fig:data_generation}
\end{figure}

\begin{figure}[h!]
    \centering
    \includegraphics[scale=0.6]{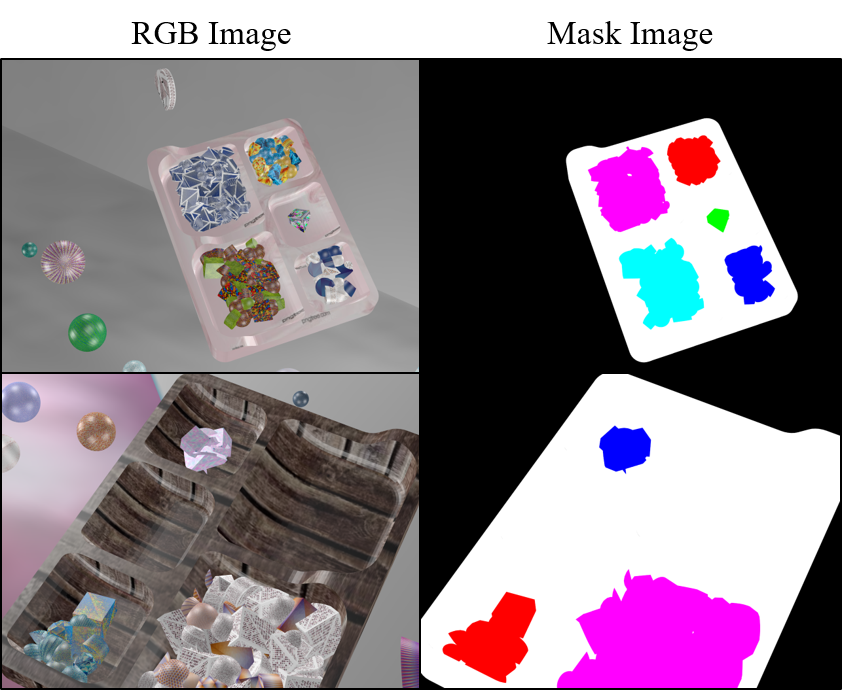}
    \caption{Examples of synthetic dataset}
    \label{fig:example_syn_data}
\end{figure}

In general, food is usually served in bowls and plates. Especially, meal tray is usually used in hospital and school. We actively introduce domain randomization to ensure that the distribution of synthetic data includes real-world data. So we created a random texture on the meal tray to recognize a variety of plates robustly, and created a variety of background textures and distractors around meal tray to be robust against environmental changes. In addition lighting conditions, such as the number, position, and intensity of light points in the virtual environment, also changed during data generation phase. To express food in synthetic data, various kinds of primitives were grouped together and placed on a plate, that resemble food with various colors and textures on the meal tray so that the network can recognize various foods robustly. Therefore, we placed meal tray modeled using the blender and generated objects of various sizes and shapes on the meal tray in a virtual simulation space as shown in Fig \ref{fig:data_generation}. We then generated synthetic data by capturing it at various angles and locations of camera. We created 28,839 synthetic data, including RGB images and mask images, for unseen food segmentation. As shown in Fig \ref{fig:example_syn_data} as the examples of dataset,  textures on the plate and background are in a complex form of mixed colors and patterns. Food-like objects located in the food tray and distractors outsider of meal tray composed of clustered primitives also have diverse colors. However, in the mask images, only the objects expressing food are projected as instance for segmentation, while the distrcators are expressed as background.

\begin{figure}[h!]
    \centering
    \includegraphics[scale=0.35]{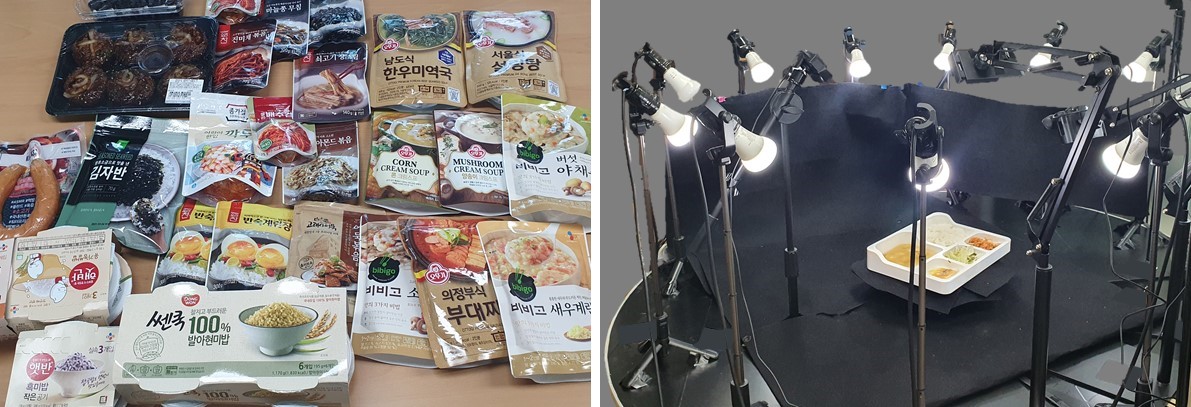}
    \caption{(Left) Examples of Korean food (Right) Data acquisition system}
    \label{fig:data_aqusit_system}
\end{figure}

\begin{figure}[h!]
    \centering
    \includegraphics[scale=0.35]{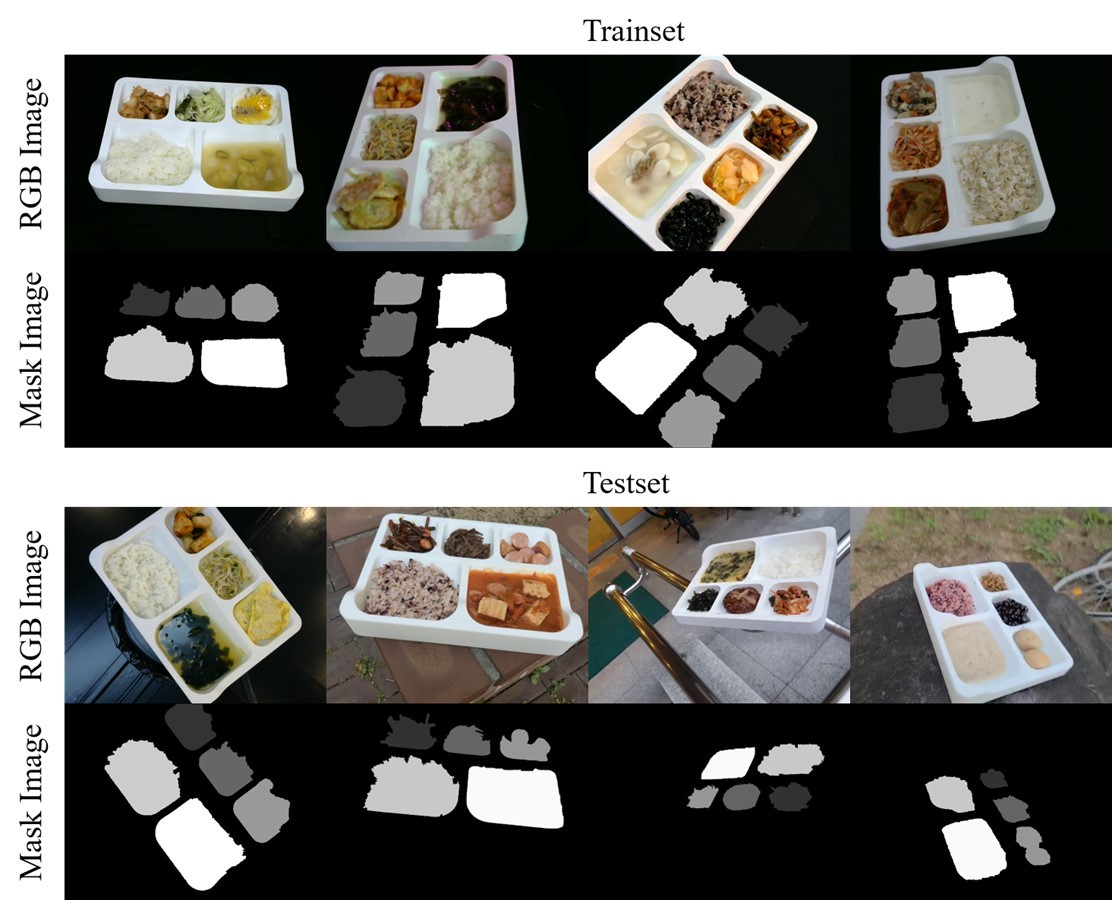}
    \caption{Examples of real-world dataset.}
    \label{fig:example_real_data}
\end{figure}

\textbf{Real-world Dataset}
We built a real-world food dataset for 50 kinds of Korean food. We selected 50 kinds of Korean food (rice, kimchi, fried egg,etc) through consultation and investigation by experts of hospital and institution, and collected dataset. Each meal tray was assembled with five food items shown in Fig \ref{fig:data_aqusit_system}. We built a real-world food dataset using the food acquisition system that captures images from various angles, heights, and lights as shown in fig \ref{fig:data_aqusit_system}. We generated data from various backgrounds to verify the robustness of the network even in environmental changes. We make a real-world food dataset by annotating 229 images acquired through the food image acquisition system. The examples of dataset is shown in  Fig \ref{fig:example_real_data}.

\subsection{Deep Neural Network}
\begin{figure*}[h!]
    \centering
    \includegraphics[scale=0.35]{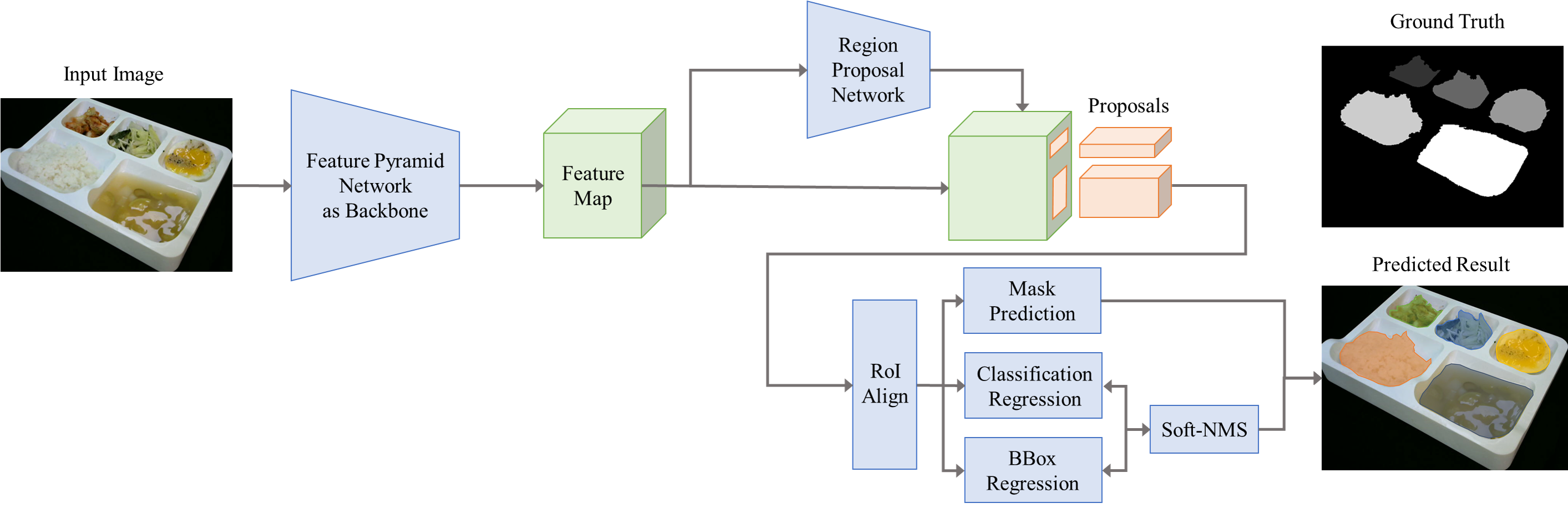}
    \caption{The architecture of Mask R-CNN using for food instance segmentation.}
    \label{fig:maskrcnn}
\end{figure*}

We used Mask R-CNN \cite{8237584} that widely used in the instance segmentation.
Mask R-CNN \cite{8237584} is an extension of the Fast RCNN \cite{girshick2015fast}, an algorithm used for object detection.
The overall network architecture are shown in Fig . As shown in the Figure \ref{fig:maskrcnn}, Mask R-CNN \cite{8237584} consists of Backbone network, region proposal network(RPN), feature pyramid network(FPN), RoIAlign, and classifier.  Mask-RCNN \cite{8237584} is built on a backbone convolutional neural network architecture for feature extraction. Backbone network used a feature pyramid network based on a ResNet-50. In feature pyramid network, the features of various layers are considered together in a pyramid-shaped manner, it gives rich semantic information compared to single networks that use only the last feature. Region proposal network is a network that scans images by sliding window and finds areas containing objects. We refer to the area that RPN searches as anchors and use RPN predictions to select higher anchors that are likely to contain objects and refine their location and size. On the last stage, Region proposal network uses the proposed ROI to perform class preference, bounding-box regression, and mask preference.  We give data and ground truth of food image as input to the network and we get output the instances of segmentation.

\subsection{Training Details}
We trained MASK R-CNN \cite{8237584} model implemented in PyTorch \cite{paszke2017automatic} with stochastic gradient descent(SGD) optimizer configured with learning rate of 0.0001, weight decay of 0.00005 and batch size of 8 on Titan RTX (24GB) GPU. We trained model on three types, first training with only synthetic dataset, second training only real-world dataset, the last fine-tuning with real-world dataset after pre-triaing on synthetic dataset. During fine-tuning the model, the model trained with only synthetic data first, and then only real dataset is used to fine-tune the pre-trained model.

\subsection{Evaluation Metrics}
For performance evaluation for unseen food segmentation, we utilize the same metric of COCO dataset \cite{lin2014microsoft}, one of the most popular criteria of instance segmentation. The Intersection over Union (IoU), also known as the Jacquard Index, is a simple and highly effective rating metric that calculates the overlapping area between the predicted and ground truth divisions: \textit{IoU=area of overlap/area of union}. The proposed outputs of segmentation are post-processed with non-max suppression by the threshold of 0.5 for IoU.

The mean Average Precision(mAP) is used evaluation metric of the performance of the instance segmentation. Precision and recall are required to calculate the mAP. Precision means the true positive ratio of predicted results which can be calculated by adding true positive and false positive: \textit{Precition=true positive/(true positive+false positive)}. Recall means the true positive ratio of all ground truths which can be calculated by adding true positive and false negative: \textit{Recall=true positive/(true positive+false negative)}. Therefore, a high Recall value means that deep neural network recorded a high proportion of the predicted results among ground truths.

This results in mean Average Precision(mAP) being obtained through the Recall and Precision values. The main metric for evaluation is mean Average Precision(mAP), which is calculated by averaging the precisions under Intersection over Union(IoU) thresholds from 0.50 to 0.95 at the step of 0.05.

\section{EXPERIMENT AND RESULTS}
We experimented that training the MASK R-CNN \cite{8237584} model on synthetic dataset and evaluation on our real-world dataset to verify the performance of unseen food segmentation. Furthermore, we conducted an experiment using a public dataset to verify the generalized performance of the algorithm. In all the experiments, our trained model segments food instances, which are category-agnostic and only certain to be food as a single category.  

\subsection{Result on our dataset} 
We categorized our real-world dataset into three types: easy, medium, and hard, based on background diversity within the image. Easy samples have a completely black background, medium samples have a black background with light reflection and hard samples have a wide variety of backgrounds. We have 73 easy samples, 61 medium samples, and 95 hard samples. Easy samples were used for training and medium and hard samples were used for testing. 

The experimental results can be found in Table \ref{tab:results_seg}. The two columns of Table \ref{tab:results_seg} (headed as \textit{Synthetic Only} and \textit{Real Only}) demonstrate the performance of models that trained only synthetic data and real data from-scratch, respectively. The column of \textit{Syn+Real} shows the performance of the model fine-tuned on real-world data after pre-training on synthetic data. The real-world data utilized on each training phase, are our dataset and public dataset UNIMIB2016 \cite{ciocca2016food2}, headed on each rows. Sim-to-Real can show good performance by training network using similar synthetic data to real-world reported on the column of \textit{Synthetic Only}. Our network trains with only synthetic data and shows 52.2\% in terms of mAP as a result of evaluating with real-world data. The result suggested that the network learned by using only synthetic data via Sim-to-Real to become unseen food segmentation for real-world data. Furthermore, we confirm that the performance increased by about 8.8\%  when the model was fine-tuned with real data compared to learning with real-world data from scratch. As shown in Figure \ref{fig:inference_seg}, the model trained with synthetic data only tends not to recognize watery foods such as soup. This seems unresponsive due to the lack of liquid modeling in training synthetic data, but it is simply overcome by fine-tuning with real data. Also the fine-tuned model shows the advantage of robustness not mistaking in the background compared to the model trained with real data only.

\begin{table}[]
\begin{threeparttable}
\renewcommand{\arraystretch}{1.2}
\renewcommand{\tabcolsep}{1.5mm}
\caption{Segmentation evaluation results of mask AP and box AP for each dataset.}
\label{tab:results_seg}
\begin{tabular}{clccc}
\hline 
Test Sets & Metric & Synthetic+Real$^{1}$ & Synthetic Only & Real Only  \\ \hline
Our test set & BBOX$^{2}$ & - & 80.0 & - \\
(Synthetic) & SEG$^{2}$  & - & 87.9 & - \\ \hline
Our test set & BBOX & \textbf{76.1} & 51.4 & 65.6 \\
(Real) & SEG & \textbf{79.0} & 52.2 & 72.6 \\ \hline
UNIMIB & BBOX & \textbf{80.6} & 35.7 & 79.3 \\
2016 & SEG & \textbf{82.7} & 32.9 & 81.7 \\ \hline
\end{tabular}
\begin{tablenotes}
\footnotesize
\item $^{1}$Synthetic+Real means pre-training with synthetic data and then fine-tuning with real-world data.
\item $^{2}$BBOX means box AP@all and SEG means mask AP@all as defined in COCO dataset \cite{lin2014microsoft}.
\end{tablenotes}
\end{threeparttable}
\end{table}

\subsection{Result on public dataset}
The UNIMIB2016 \cite{ciocca2016food2} has been collected in a real canteen environment. The images contain food on their plates and are also placed outside their plates. In some cases, there are several foods on a plate. The UNIMIB2016 is a dataset for food instance segmentation that captures food from the top view. The UNIMIB2016 \cite{ciocca2016food2} is composed of 1,010 tray images with multiple foods and containing 73 food categories. The 1,010 tray images are split into a training set and a test set to contain about 70\% and 30\% of each food instance, resulted in 650 tray image training sets and 360 image test sets. Although the UNIMIB2016 \cite{ciocca2016food2} contains the categories of each food, we utilize all data as single category, food, for comparison with our unseen food segmentation performance. 

\begin{figure}[h!]
    \centering
    \includegraphics[scale=0.35]{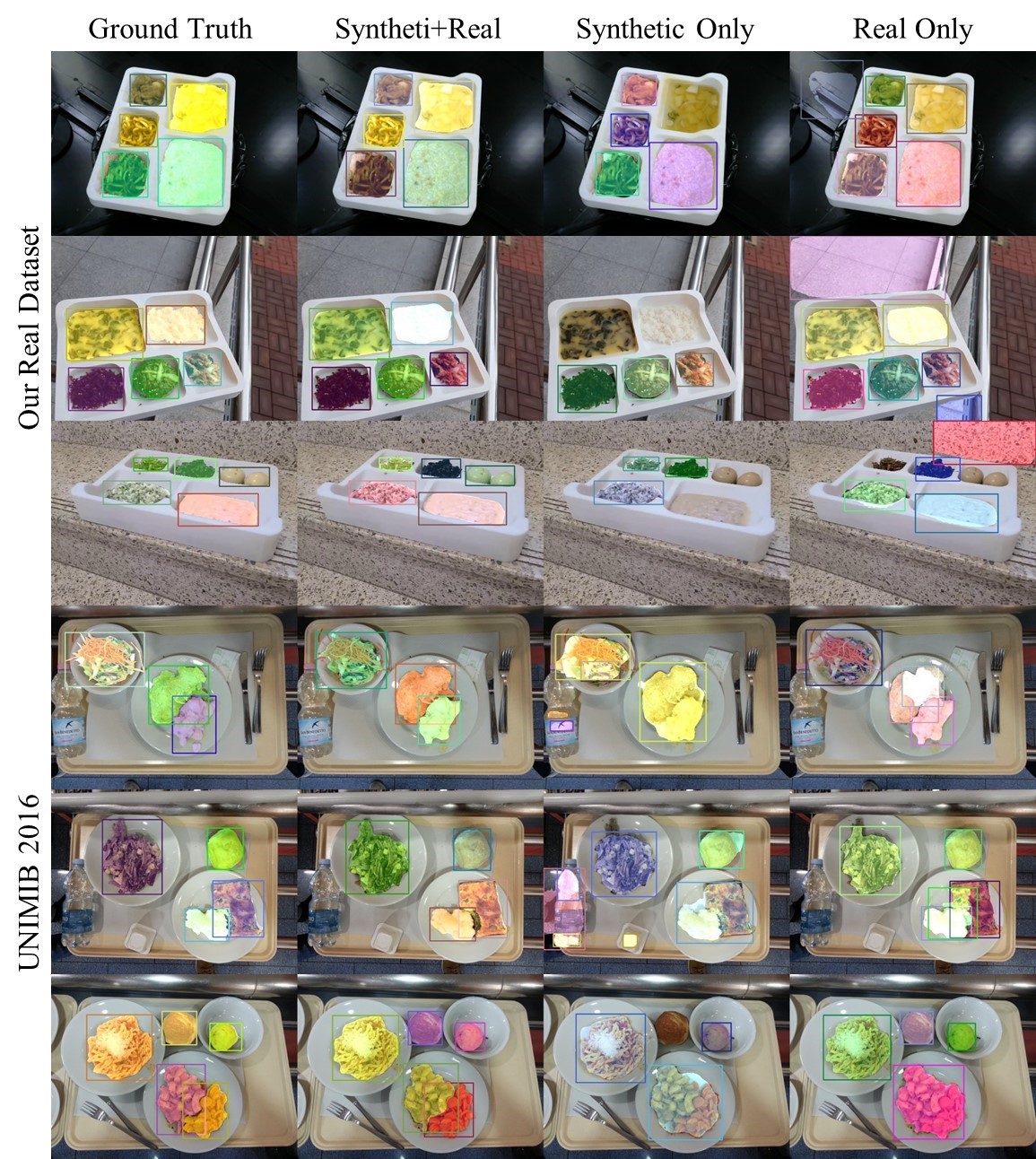}
    \caption{Inference examples of segmentation results}
    \label{fig:inference_seg}
\end{figure}

We conducted experiments using synthetic data, UNIMIB2016 \cite{ciocca2016food2} as real-world data, fine-tuning with real-world data after pre-training on synthetic data, and the results can be seen through Table \ref{tab:results_seg}. When the network was trained with only the synthetic data, mAP was 32.9. Because some data of UNIMIB2016 \cite{ciocca2016food2} dataset is several food closely attached on a same plate,   Although the network did not train with foods in the UNIMIB2016 \cite{ciocca2016food2}, network can implement food instance segmentation as shown in Figure \ref{fig:inference_seg}. Unlike synthetic data, because some data in the UNIMIB2016 \cite{ciocca2016food2} dataset multiple foods are clustered together on the same plate, the model trained on synthetic data tends to recognize foods on a single plate as one instance. Despite, using real-world data shows better results than using the synthetic data, in the case of training with fine-tuning with real-world data after pre-training on synthetic dataset, the highest result was obtained with 82.7\% in terms of mAP. As a result, training on synthetic dataset is applicable to real-world data via Sim-to-Real and also takes a roll of general feature extraction that is more appropriate for fine-tuning as task-specific adaption.

\section{CONCLUSIONS}
In this paper, we demonstrate the possibility of food instance segmentation that have never been seen in real-world environment through synthetic data generation and training of Mask R-CNN \cite{8237584} model. On our real-world dataset, food instances can be segmented sufficiently with a performance of 52.2\% as using a network learned from only synthetic data. Also, when fine-tuning a model learned from only synthetic data with real-world data, +6.4\%p performance is improved better than the model trained from scratch. Experiments on public dataset(UNIMIB 2016\cite{ciocca2016food2}) show that it is sufficient to segment food, even if it is not the same meal tray. Since this work can distinguish between different food instances but cannot recognize the type of food, it is also remaining challenge to expand intelligence for recognition of food categories. We suggest a study as our future work, transferring knowledge from classification intelligence that can be implemented with relatively easy to collect data to recognize the category of mask instance in our food instance segmentation models.

\bibliographystyle{ieeetr}
\bibliography{biblist}

\addtolength{\textheight}{-12cm}   




\end{document}